\documentclass[letterpaper, 10 pt, conference]{ieeeconf}  
\pdfoutput=1
\IEEEoverridecommandlockouts                              

\usepackage{graphics} 
\usepackage{epsfig} 
\usepackage{graphicx}
\usepackage{tabularx}
\usepackage{multirow}
\usepackage{multicol}
\usepackage{enumerate}
\usepackage{url}
\usepackage{subcaption}
\usepackage{dblfloatfix}
\usepackage{amsmath} 

\title{\LARGE \bf
Human-Aware Navigation Planner for Diverse Human-Robot Contexts
}

\author{Phani Teja Singamaneni$^{1}$, Anthony Favier$^{1,2}$, Rachid Alami$^{1,2}$ 
\thanks{$^{1}$Authors are with LAAS-CNRS, Universite de Toulouse, CNRS, Toulouse, France, {\{ptsingaman, anthony.favier, rachid.alami\}@laas.fr} }%
 \thanks{$^{2}$Authors are with Artificial and Natural Intelligence Toulouse Institute (ANITI)}%
}

\begin{document}

\maketitle
\thispagestyle{empty}
\pagestyle{empty}

\begin{abstract}
As more robots are being deployed into human environments, a human-aware navigation planner needs to handle multiple contexts that occur in indoor and outdoor environments. In this paper, we propose a tunable human-aware robot navigation planner that can handle a variety of human-robot contexts. We present the architecture of the planner and discuss the features and some implementation details. Then we present a detailed analysis of various simulated human-robot contexts using the proposed planner along with some quantitative results. Finally, we show the results in a real-world scenario after deploying our system on a real robot.  
\end{abstract}

\section{Introduction}
In the recent decade, more and more robots are being deployed into human environments. From the robotic vacuum cleaners to the human assisting robots in shops, malls \cite{kanda2009affective, foster2019mummer} and airports \cite{wettergreen_spencer:_2016}, all of these robots are working in environments with humans moving around. To navigate in these places, the robot needs to be aware of the humans in the environment, and treating humans simply as obstacles may not be enough. Besides, the robot's motion should be safe, legible and acceptable to the humans rather than being optimal. Therefore, a new field of robotic navigation called the human-aware (or social) navigation has emerged, which studies various human motion and social aspects for developing more human acceptable robotic navigation. 

Human-aware navigation can be broadly divided into two categories based on the environmental context, 1) Crowd or outdoor navigation and 2) Indoor navigation. In the context of crowd navigation, the planner needs to be reactive and plan its motion quickly to avoid any possible collisions or obstructions to people. On the contrary, in the indoor contexts, a highly reactive planner is not preferable as the robot’s actions may cause confusion \cite{kruse_evaluating_2014} among the humans in the shared space or lead to navigation failure in various intricate scenarios. Therefore, a different set of navigation planners are developed for various types of indoor environments with shared human spaces like warehouses  \cite{althoefer_making_2019}, offices \cite{truong_dynamic_2014}, labs, homes \cite{kollmitz_time_2015} etc. This is because there is no single algorithm that can cover all environments and situations. However, changing the navigation algorithm every time to address a new human-robot context may not be ideal.  In order to address these issues, we propose a highly tunable human-aware navigation system with multiple modes of planning that can be employed in a variety of human-robot contexts, simply by adjusting the system parameters. Our main contributions in this work are threefold and are summarized below:
\begin{enumerate}
    \item We propose a tunable human-aware navigation planner with different planning modes that can handle very complex indoor scenarios as well as crowded scenarios.
    \item We extend our previous work, Human-Aware Timed Elastic Band (HATEB) \cite{singamaneni2020hateb} local planner, to effectively handle large numbers of people and to offer more legible and acceptable navigation.
    \item We evaluate the proposed planner in several simulated human-robot scenarios and present both qualitative and quantitative analysis. Further, we also present some tests conducted on the real robot at our lab.
\end{enumerate}
The rest of the paper is organized as follows. 
In section \ref{rel_work}, we discuss the related works, and in section \ref{sec_2} planner's architecture is presented along with explanations of various modules and features. Following this, section \ref{sec_3} presents the evaluation of our planner in various simulated human contexts. We also present a comparison with one of the existing human-aware navigation planners. In section \ref{sec_4}, we talk about the tests conducted on the real robot. Finally, section \ref{sec_5} presents some discussions and conclusions.
\section{Related Work}\label{rel_work}
There are a variety of human-aware navigation planners designed for different human-robot contexts. In the crowd context, Ferrer. et al \cite{ferrer_social-aware_2013} presents a potential field based navigation using the social force model. The authors of \cite{truong_toward_2017} extended this to human-object and human group interactions by proposing the proactive social motion model. The work by Repiso et. al \cite{repiso2017line} shows the context of a robot accompanying a human. The authors of \cite{chen_socially_2017} address this crowd navigation problem by using reinforcement learning. Coming to the indoor contexts, the works presented in \cite{sisbot2007human,truong_dynamic_2014} and \cite{kollmitz_time_2015} show some interesting costmap based approaches for planning paths in complex indoor scenarios that can occur at homes or offices. In this paper, we use a similar costmap based approach to handle static humans. Fernandez Carmona et. al \cite{althoefer_making_2019} compares the performance of the existing navigation planners in a warehouse context and proposes an architecture to include humans into planning. The work of G\"{u}ldenring et. al \cite{guldenring2020learning} addresses the same context using reinforcement learning. Khambhaita and Alami \cite{khambhaita2017viewing} addressed the context of human-robot co-navigation based on an optimization-based approach. Note that none of the above planners was designed to handle multiple human contexts together. 

A multi-context human-aware navigation planning is a very new field, and not many works exist. The work by Banisetty et. al \cite{banisetty2020deep} shows some preliminary results using a deep learning-based context classification and multi-objective optimization based navigation planner. However, these results are in indoor scenarios with static humans, and the authors do not present any results in dynamic human scenarios, whereas the human-aware navigation planner proposed in this paper can handle static as well as dynamic humans in both indoor and crowd contexts. 

In order to handle the dynamic humans in our navigation planner and to plan a socially acceptable trajectory for the robot, a human motion prediction system is required. One of the classic approaches of human motion prediction is based on the social force model \cite{helbing1995social}. Ferrer et. al \cite{ferrer2015multi} uses this social force model both to predict human motions and to move the robot among the crowds. Kollmitz et. al \cite{kollmitz_time_2015} uses a simple linear prediction based on current human velocity. Instead of predicting the trajectory, a possible human goal can also be predicted using certain reasoning over a probable set of goals \cite{bordallo2015counterfactual}. Our proposed navigation system uses one such goal prediction system \cite{ferrer2014bayesian} as a part of the human path prediction module. Apart from this, our system offers three other human path prediction methods to handle a variety of situations. In a recent work by Fisac et. al \cite{fisac2018probabilistically}, the authors suggest a probabilistic human model with confidence to handle the uncertainties in a system.

One of the key elements of the proposed human-aware navigation planner is the context-based shifting between different planning modes. This kind of modality shifting is discussed in the works of Qian et. al \cite{qian_decision-theoretical_2013}, and Mehta et. al \cite{mehta_autonomous_2016} based on Partially Observable Markov Decision Processes. Unlike these, our system uses situation assessment based modality shifting. In our previous work, \cite{singamaneni2020hateb}, we introduced this modality shifting with three different modes of planning. In the current work, we extend this to handle a large number of humans and also introduce some elements including a new planning mode. This modified \textit{HATEB local planner} is integrated into the proposed human-aware navigation framework as the local planner. 

\section{Planner Architecture and Features} \label{sec_2}
\begin{figure}[h!]
    \centering
    \includegraphics[width=0.8\linewidth]{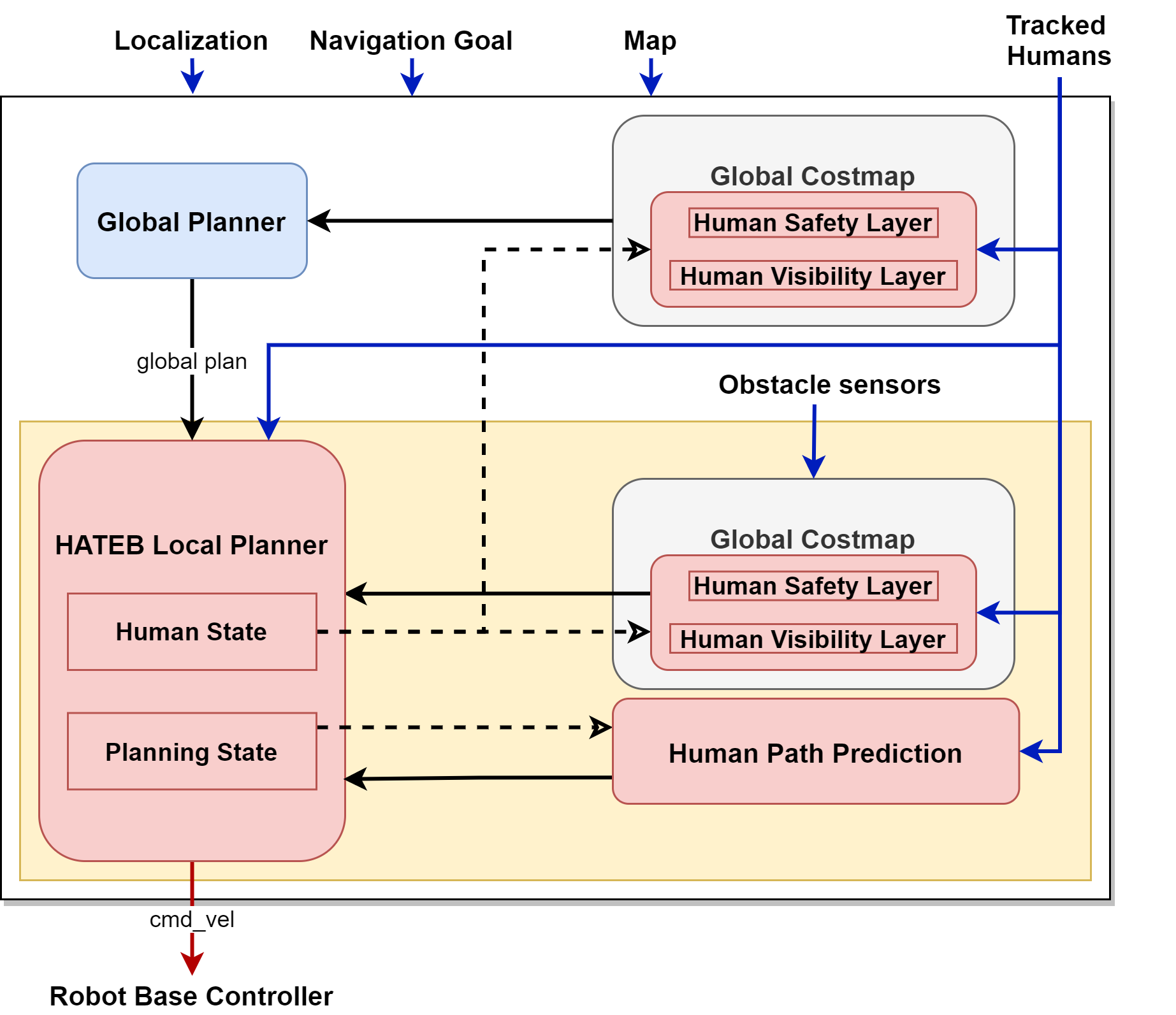}
    \caption{Software architecture of the proposed planner.}
    \label{fig:block_diag}
\end{figure}

In this section, we present the overall architecture of the human-aware navigation planning system and explain its features that allow us to deal with various kinds of human-robot contexts. Our system is developed over the ROS \cite{quigley2009ros} navigation stack, and the architecture of the proposed system is shown in Fig. \ref{fig:block_diag}. 
The red blocks shown in the Fig. \ref{fig:block_diag} are the modifications we introduced into the standard ROS navigation stack and are the major contributions of this work. As shown in the figure, we introduce \textit{Human Safety} and \textit{Human Visibility} costmaps layers into both global and local costmaps.  The \textit{Human Safety} layer is modelled as a 2D Gaussian around the human, and the \textit{Human Visibility} layer as a 2D half Gaussian on the backside of the human. Both these layers have a cutoff radius of $3m$ \cite{sisbot2007human} beyond which the cost is zero.  These layers are implemented using a costmap plugin that we developed for ROS called the \textit{human\_layers}\footnote{{\url{https://github.com/sphanit/human_layers/tree/tested}}}. The addition of these layers is controlled by the \textit{Human State} of the \textit{HATEB local planner}. The \textit{Human Path Prediction}\footnote{{\url{https://github.com/sphanit/hanp_prediction/tree/tested}}} module predicts the possible paths for the requested humans using the selected prediction method. The \textit{HATEB local planner}\footnote{{\url{ https://github.com/sphanit/hateb_local_planner/tree/tested}}} module accesses different human-robot scenarios and determines the \textit{Human State} and the \textit{Planning State} shown in the figure. Both these states together decide the planning mode of the system and also control the transition between different modes. Based on the \textit{Planning State}, the appropriate path prediction method is selected for the humans. After accepting a navigation goal, the system continuously accesses the human-robot scenario and appropriately chooses a planning mode that decides the command velocity sent to the robot's base controller. Note that the planning mode need not be constant and can shift depending on the context. Further, our system is completely tunable, and the transition between different modes can be tuned (or changed) by changing the mode transition parameters \cite{singamaneni2020hateb} as per the requirement.
\subsection{Types of Humans and Costmap layers}
In our system, we deal with different types of humans while navigating the robot to the goal. 
\begin{figure}[h!]
    \centering
    \includegraphics[width=0.7\linewidth]{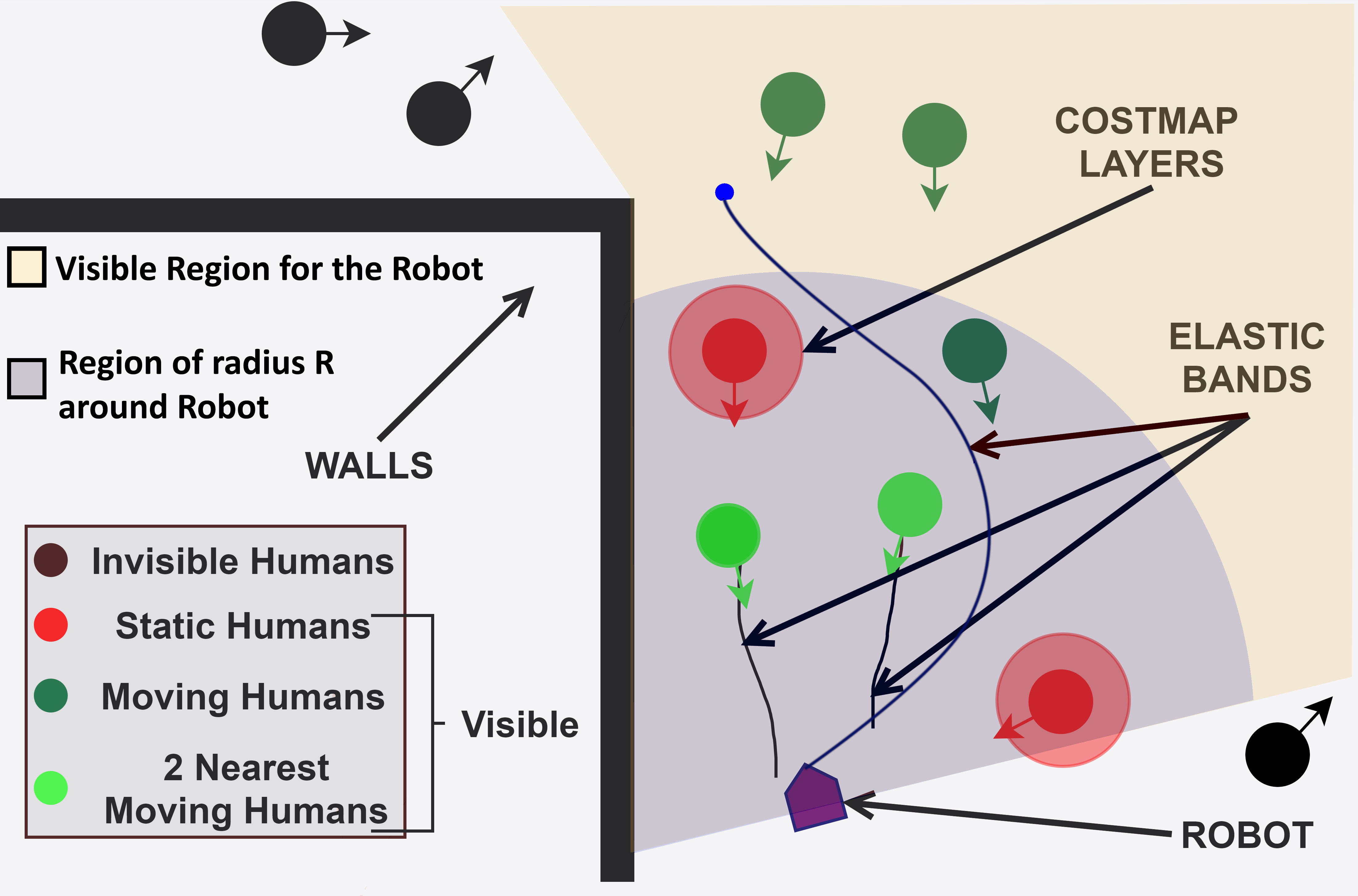}
    \caption{Different types of humans considered in our system. A sample trajectory of the robot is shown among different humans.}
    \label{fig:human_types}
\end{figure}
Fig. \ref{fig:human_types} shows all these types of humans along with the robot's visibility and the planning radius, R. While the robot is moving in the environment, the system considers only the humans within this planning radius that are in the visible region. Among these humans, it checks for the static and dynamic humans and updates the \textit{Human State} for all the humans.
The \textit{human\_layers} plugin checks the \textit{Human State} of all the observable humans and adds the \textit{Human Safety} and \textit{Human Visibility} costmap layers around the static humans. We chose to add these costmap layers only around static humans because the response time for static humans is usually slow, and hence the robot should maintain a larger safety distance as well as avoid surprise appearances from behind \cite{sisbot2007human}. Besides, as no elastic band is added to the static humans, the addition of these layers is necessary to plan a safe path. For the dynamic humans within the planning radius, the system adds elastic bands to the nearest two humans and their path, as well as trajectories, are predicted until they move behind the robot or out of the planning radius.
\subsection{Human Path Prediction}
The \textit{Human Path Prediction} module deals with different kinds of human goal predictions and building the global plans for required humans. Our system currently offers four types of human goal prediction and path planning methods.
\begin{enumerate}
    \item \textit{PredictBehind}: This method predicts that the human goal is behind the robot. The position of the robot when the human enters the visible planning radius is used for this. This goal is used to predict the path.
    \item \textit{PredictGoal}: This method predicts the most probable goal among the set of goals provided to the system using the approach described in \cite{ferrer2014bayesian}. The predicted goal is then used for path prediction.
    \item \textit{PredictVelObs}: This method builds a path by extrapolating the current human velocity over a fixed duration and does not predict any goal. Currently, the duration is set to $5s$. This is the default prediction service in \textit{VelObs} mode.
    \item \textit{PredictExternal}: This service accepts a goal from an external system and adds a global path prediction based on the provided goal.
\end{enumerate}
These services provide global plans for the humans that are used by the \textit{HATEB local planner} for planning local plans (or trajectories). 

\subsection{HATEB local planner and planning modes}
This is the core module of the proposed human-aware navigation planning system. \textit{HATEB local planner} is based on the human-aware extension of Timed Elastic Band (TEB) \cite{rosmann2013efficient} local planner by Khambhaita and Alami \cite{khambhaita2017viewing}. This module plans the robot's trajectory, as well as the possible human trajectories for the nearest two humans in the visible planning radius, based on the predicted human paths. It continuously assesses the current human-robot context and sets the \textit{Planning State} and the \textit{Human State}. Depending on the value of these states, it shifts between different planning modes. This is needed in the intricate human-robot contexts that cannot be solved using a single planning mode.\\
\textbf{\textit{{Modes of Planning}}}: \textit{HATEB local planner} provides four different modes of planning at the control level and intelligently shifts between them based on the situation. \\
1. \textit{{SingleBand mode}}: This is the mode in which the planning system starts and has an elastic band only for the robot. The system stays in this mode as long as there are no humans within the visible planning radius. The default planning radius is chosen as $10m$.\\ 
2. \textit{{DualBand mode}}: In this mode, elastic bands are added to the two nearest dynamic humans in visible planning radius and trajectories are planned for humans along with the robot. This kind of human planning allows the robot to proactively plan its trajectory and adapt to the changing human plans. On top of being useful as predictions, these planned trajectories also offer a possible solution for the human-robot context, if followed, will resolve any conflict that exists.\\
3. \textit{{VelObs mode}}: This mode uses all the human-aware criteria while planning, but adds bands to humans only if they have some velocity.
This mode is useful in crowded human scenarios or when the robot cannot move due to entanglement issues \cite{singamaneni2020hateb} of the \textit{DualBand} mode.\\
The situation assessment and mode shifting scheme for the above-mentioned planning modes is described clearly in our previous work \cite{singamaneni2020hateb}. In this work, we extended this by adding a \textit{Backoff-recovery} mode.\\ 
4. \textit{{Backoff-recovery mode}}: The \textit{Backoff-recovery} mode is required when there is no solution to the planning problem unless one of the agents completely clears the way for the other. This kind of situation commonly occurs in a very narrow corridor where only one person (or robot) can navigate at the same time. If a human and robot face each other in a very narrow corridor or another situation where one of them has to clear the way for the other, our system gives priority to the human and makes the robot clear way for the human. This is done by making the robot move back slowly until it can go either left or right to clear the way. Once the robot clears the way, it waits for the corresponding human to complete its navigation or a timeout and then proceeds to its goal. This mode is activated when the robot is in \textit{VelObs} mode in the near vicinity of the human ($<2.5m$), and it is stuck without progressing towards the goal for more than 10 seconds. It can also accept new goals in the waiting state and navigate to them, discarding the existing goal. 

\textit{HATEB} uses several human-aware constraints in its optimization scheme for proactive and legible planning around humans in the environment. 
Several of these constraints are listed in our previous works. In this work, we have added two new constraints along with the previous constraints present in the system. These new constraints, called the \textit{Visibility} and \textit{Relative Velocity} are explained briefly below:

\textit{Visibility}: This constraint adds cost to the optimization when the robot is behind the human and it plans to cross the human or to go in front of the human. This tries to avoid the emerging of the robot suddenly from behind the human and to emerge into the human's field of view from a larger distance. It is implemented by adding a 2D half Gaussian behind the human.

\textit{Relative Velocity}: This constraint adds cost to the optimization based on the relative velocity between humans and the robot and their distance. The main effect of this constraint is low robot velocity in the close vicinity \cite{kruse_evaluating_2014} of the human if it cannot find a path to maintain a greater distance. If the robot can find a path with a greater distance from a human, it chooses that path with a normal velocity profile. Another effect of this constraint is early intention show of the robot similar to \textit{TTC} or \textit{TTCplus} constraint given in \cite{singamaneni2020hateb}. The cost added is shown below.
\begin{equation}
    cost_{rel\_vel}=\frac{((max(\overrightarrow{V_{rel}}\cdot\overrightarrow{P_r P_h}),0) + \lVert\overrightarrow{V_r}\rVert + 1)}{\lVert\overrightarrow{P_r P_h}\rVert}
\end{equation}
where $\overrightarrow{P_r},\overrightarrow{V_r}$ are the position and velocity of the robot, $\overrightarrow{P_h},\overrightarrow{V_h}$ are the position and velocity of the human and $\overrightarrow{V_{rel}} = \overrightarrow{V_r}-\overrightarrow{V_h}$. Since this constraint has similar effects as \textit{TTC} or \textit{TTCplus}, we activated only this constraint and deactivated \textit{TTC} and \textit{TTCplus} constraints in all the experiments presented in this paper. \textit{HATEB} takes all the activated human-robot constraints and other necessary kinodynamic constraints and plans the trajectories of the robot and the humans. Since the local planner runs a computationally expensive optimization in each control loop, extending the planning beyond two humans does not yield real-time control of the robot. Hence we restricted our human planning to the two nearest humans. 

One more extension in this work is the addition of the field of vision of the robot into the planning system. This is done by using ray-tracing from the robot's position on the environmental map. Since human tracking is provided by an external system, it is important to restrict the system to consider the humans present in its field of view. This is more natural and makes it easier to use our system with vision-based human-tracking. The proposed system also offers a variety of parameter settings that can choose prediction mode, the human-aware constraints to be used and tuning over these costs. Even the planning can be restricted to only one of the three planning modes (except \textit{Backoff-recovery}). Hence, it is possible to extend it to many kinds of human-robot contexts by properly choosing the parameters and with simple tuning. With the addition of the costmap layers around the static humans, this framework can handle all the scenarios presented in \cite{banisetty2020deep}. We present our results in different scenarios and environments in the next section.
\section{Results in various Simulated Scenarios} \label{sec_3}
To validate our system, we applied it to various kind of human-robot contexts that can occur in day to day life. These situations are generated in a simulated environment based on MORSE \cite{echeverria2011modular}. The humans in these simulations are controlled in three different ways to test the robustness of the system: (1) Joystick based control by a human operator, (2) Using an improved human motion simulator we have developed in our lab (under review (submitted to IROS2021)) and (3) Using the human trajectories generated by \textit{HATEB local planner} (an ideal situation where human moves exactly as the robot expects). We present in detail some of these intricate scenarios in this section, along with some quantitative results. Further, we also present some details about the extension of our system to crowded scenarios using PedSim ROS\footnote{\url{https://github.com/srl-freiburg/pedsim_ros}}. In all figures shown below, the trajectories of the robot and humans (if shown) are shown as coloured circles. These circles are the poses planned by \textit{HATEB local planner}, and the colour corresponds to the time instant. If the predicted human pose and the robot pose has the same colour, they belong to the same time instant.

\subsection{Door Crossing Scenario}
\begin{figure}[h!]
\begin{subfigure}{.5\columnwidth}
  \includegraphics[width=\textwidth]{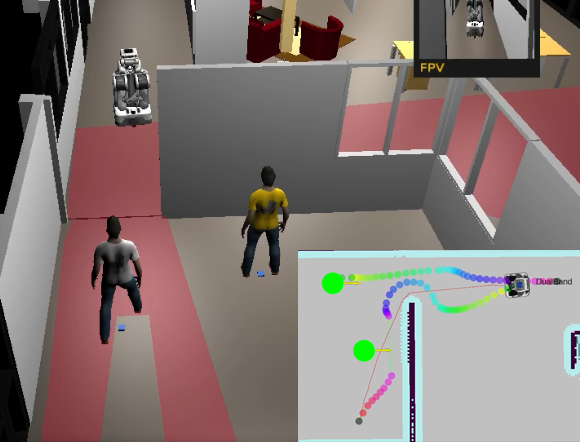}
\end{subfigure}
\hspace{-0.17cm}
\begin{subfigure}{.5\columnwidth}
  \includegraphics[width=\textwidth]{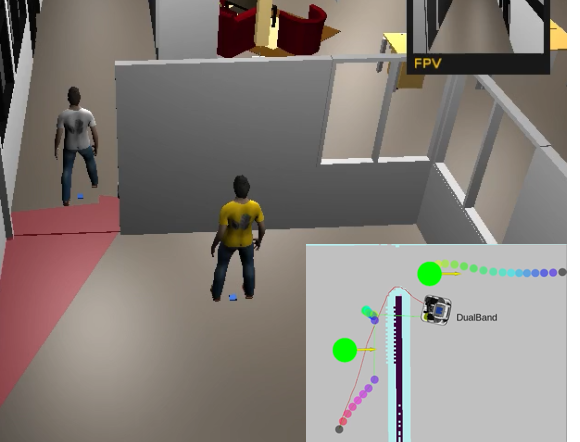} 
\end{subfigure}
\caption{Door crossing scenario in the simulated environment. The human moves towards the door. The robot sees the human and waits on the side of the door (right) until the human crosses.}
\label{fig:door_cross_scene}
\end{figure}

\begin{figure}[ht!]
\begin{subfigure}{.5\columnwidth}
  \includegraphics[width=\textwidth]{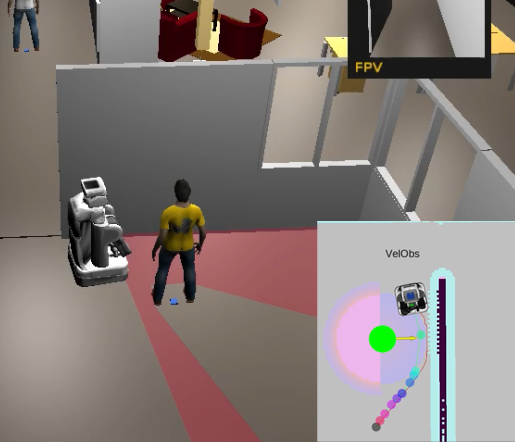}
\end{subfigure}
\hspace{-0.17cm}
\begin{subfigure}{.5\columnwidth}
  \includegraphics[width=\textwidth]{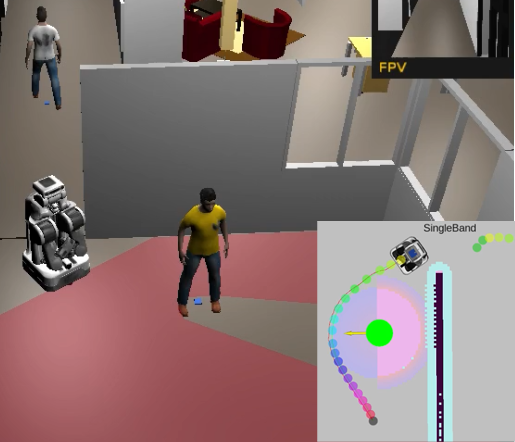} 
\end{subfigure}
\begin{subfigure}{.5\columnwidth}
  \includegraphics[width=\textwidth]{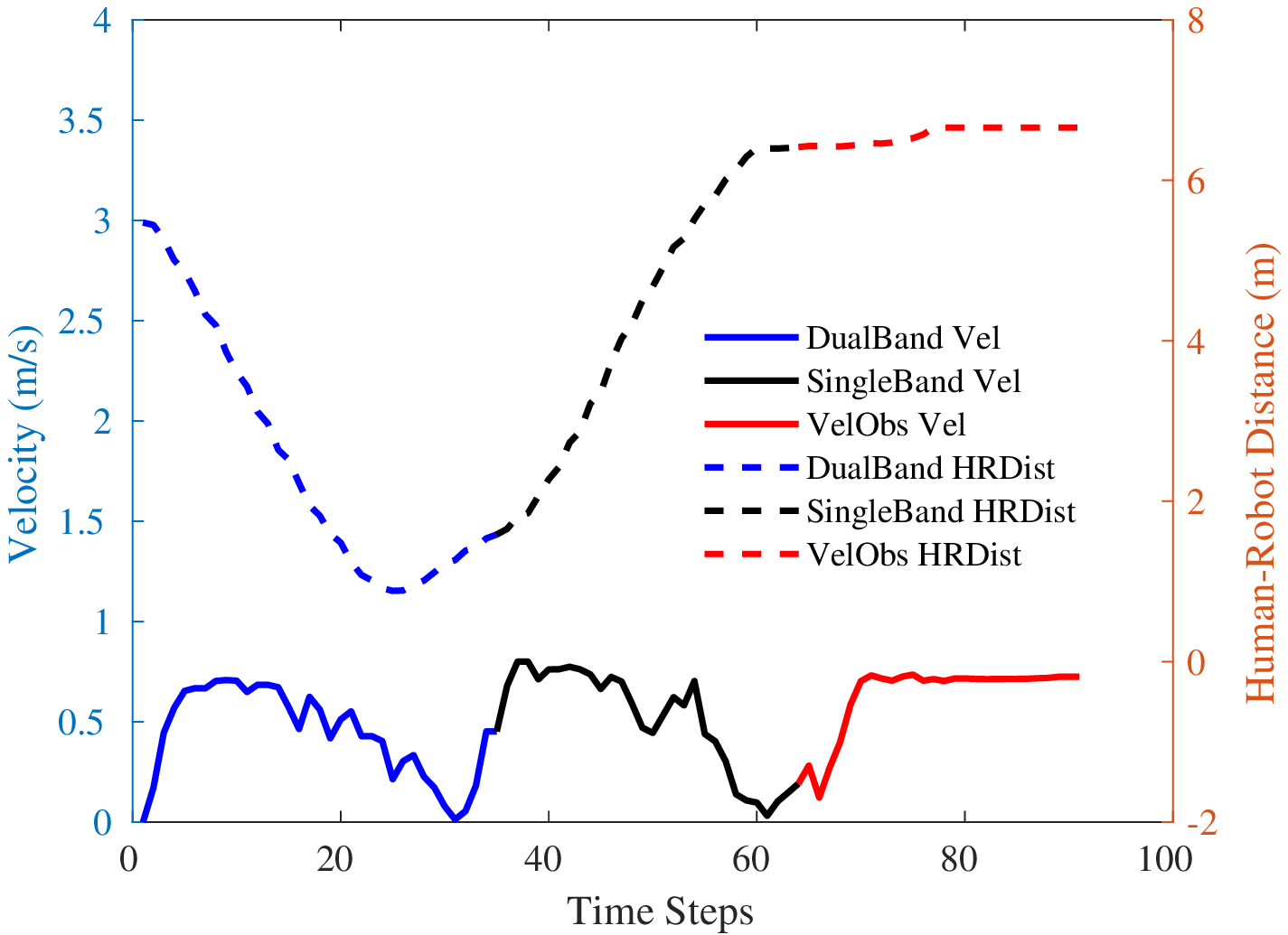}
\end{subfigure}
\hspace{-0.17cm}
\begin{subfigure}{.5\columnwidth}
  \includegraphics[width=\textwidth]{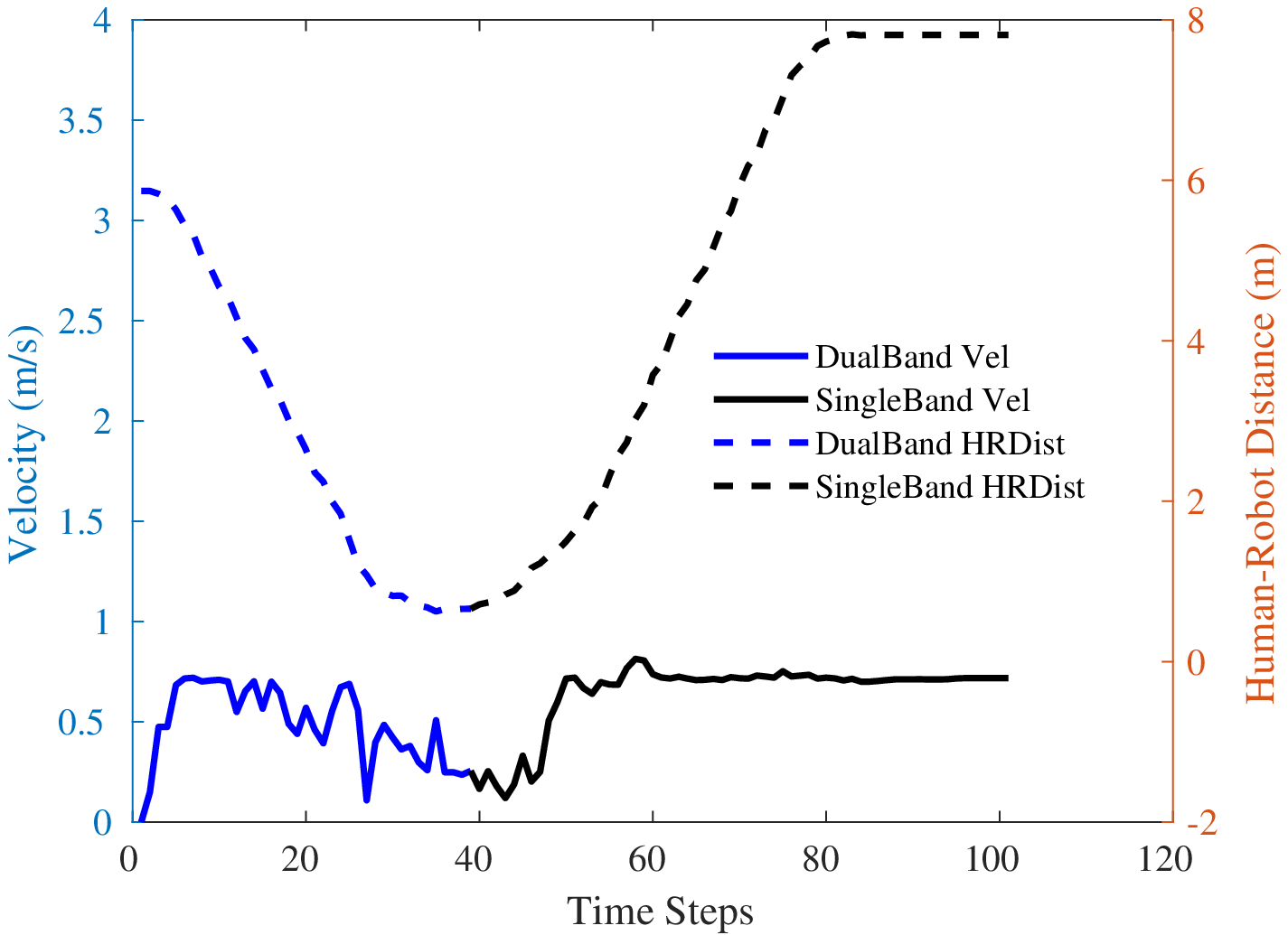} 
\end{subfigure}
\caption{Door crossing scenario in the simulated environment with the static human in two different orientations. Top two pictures show the scenarios in simulation and the planned trajectory of the robot. The bottom two figures show the robot velocity and human-robot distance graphs over time steps.}
\label{fig:door_cross_static}
\end{figure}

Door crossing is a very common and simple situation in many human environments. If two humans try to pass through the same door, one of them has to compromise and clear the way for the other. We have placed the robot running our planning system in the door crossing situation shown in Fig. \ref{fig:door_cross_scene}. The goal of the robot is beside the second human standing in the room, and the system uses \textit{PredictBehind} human path prediction. The left part of the figure shows the simulated scenario and the corresponding trajectories planned by the system for the human and the robot. The simulated human crossing the door was controlled by a human and hence does not move as the planning system expects. The system quickly adapts to these changes and makes the robot clear the way for the human by waiting on the side, as shown in the right part of Fig. \ref{fig:door_cross_scene}. The robot continues to its goal after the human crosses the robot. The planning model is \textit{DualBand} until the human crosses the robot and then it switches to \textit{SingleBand}. 
As soon as the robot crosses the door, it faces one more human, but this human is just standing at the same place and does not move. Since the human is static, our system adds the \textit{human\_layers} to the costmaps and re-plans its path. The same scenario is repeated with the second human placed in two different orientations and as shown in Fig. \ref{fig:door_cross_static}. In both scenarios, there is enough space between the wall and the human for the robot to reach its goal, maintaining a safe distance from the human. On the top left scenario of the figure, the human can see the robot, and so the planner makes the robot pass through this space. However, in the second scenario, the human cannot see the robot. Therefore, our planner completely re-plans the path as shown (top right) and makes the robot reach its goal by taking a longer path that is visible to the human. This is due to the added \textit{Human Visibility} layer.
Fig. \ref{fig:door_cross_static} also shows the plots corresponding to robot velocity (on left y-axis) and the distance between the moving human and the robot (human-robot distance) (on right y-axis) with respect to the time steps (on x-axis). Different colours in different portions of the plots correspond to different planning modes of the system, as indicated in the plots. The solid line represents the robot's velocity (Vel), while the dashed line shows the human-robot distance (HRDist). The same conventions are followed across this paper. From both the graphs (Fig. \ref{fig:door_cross_static} bottom left and right), it can be observed that the robot's velocity decreases as the human-robot distance decreases. This is a combined effect of several human-aware constraints of our system. However, the \textit{Relative Velocity} constraint plays a major role here. Secondly, it can be seen in the graph of the first scenario (bottom left) that the robot's velocity decreases one more time before the planner changes to \textit{VelObs} mode. This is because the robot is trying to navigate a narrow space between the human and the wall. This causes the planner to slowdown its velocity and checks the state of the human. Since the human is static, it shifts to \textit{VelObs} mode that has little reduced constraints and continues its navigation.
\subsection{Narrow Corridor Scenario}
\begin{figure}[ht!]
    \centering
    \includegraphics[width=0.8\linewidth]{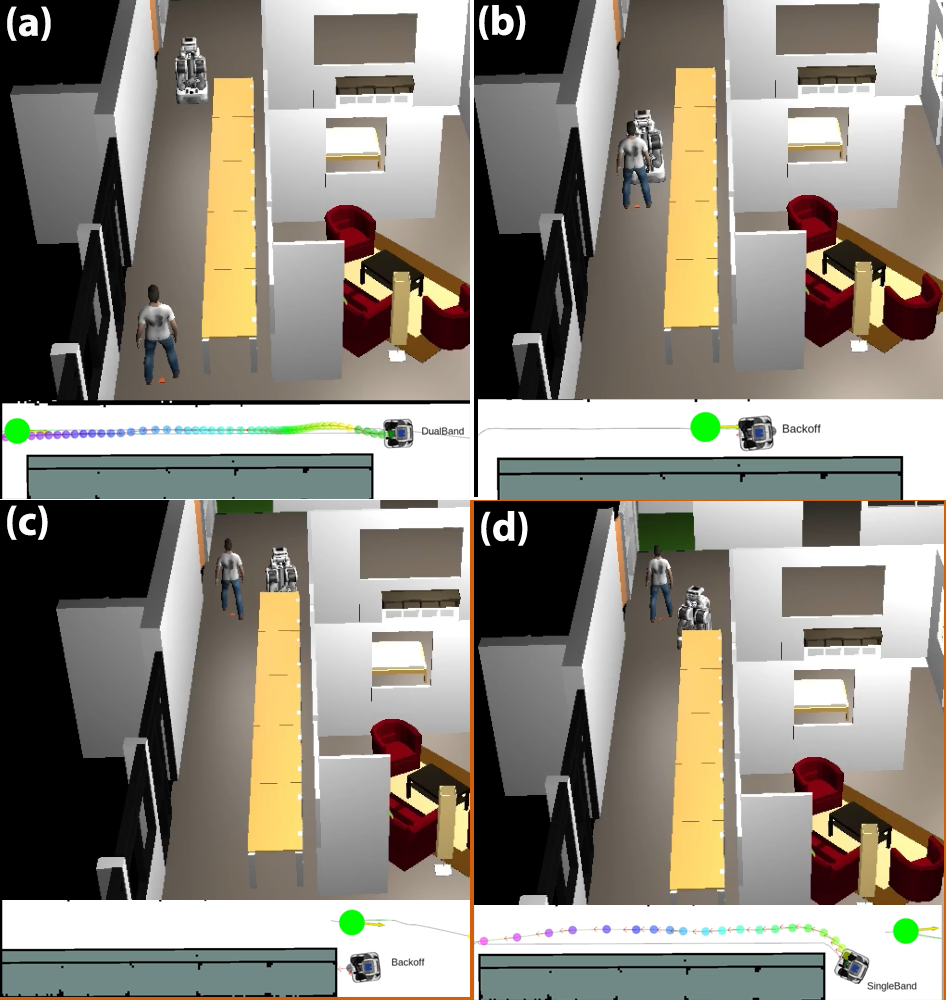}
    \caption{Narrow corridor scenario simulated in MORSE. (a) The initial planned trajectory of the robot in \textit{DualBand} mode. (b) The robot's way is blocked by the human and the system shits to the \textit{Backoff-recovery} mode. (c) The robot clears the way for the human and waits on the side until human crosses the robot. (d) The robot continues to its goal in \textit{SingleBand} mode.} 
    \label{fig:narrow}
\end{figure}
\begin{figure}[h!]
    \centering
    \includegraphics[width=0.8\linewidth]{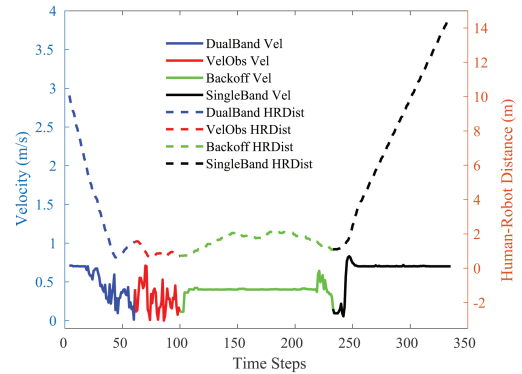}
    \caption{Plots of velocity and human-robot distance over time steps in the \textit{Narrow Corridor} scenario.}
    \label{fig:narrow_plot}
    \vspace{-0.3cm}
\end{figure}
\begin{figure*}[ht!]
    \begin{subfigure}{\textwidth}
      \includegraphics[width=\textwidth]{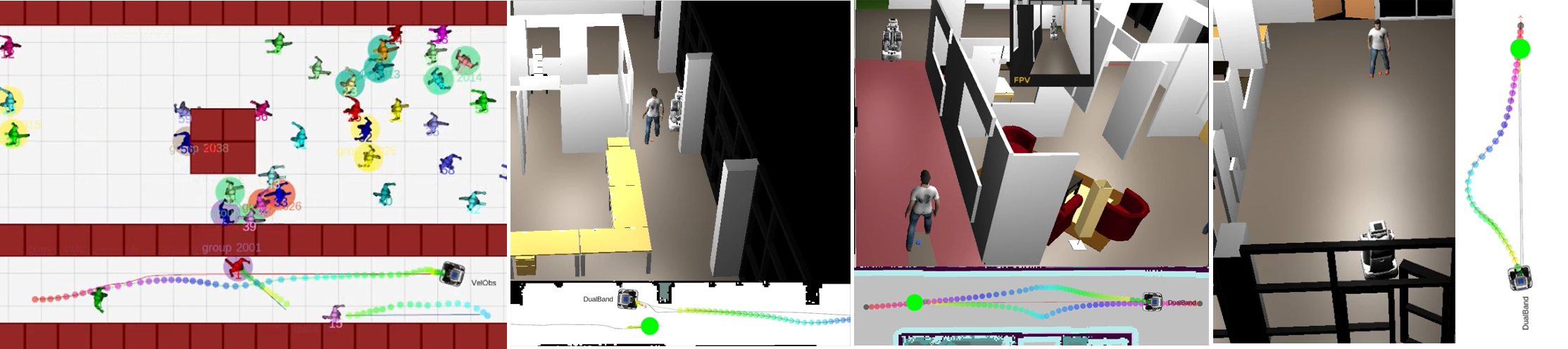}
    \end{subfigure}
    \begin{subfigure}{0.667\columnwidth}
      \includegraphics[width=\textwidth]{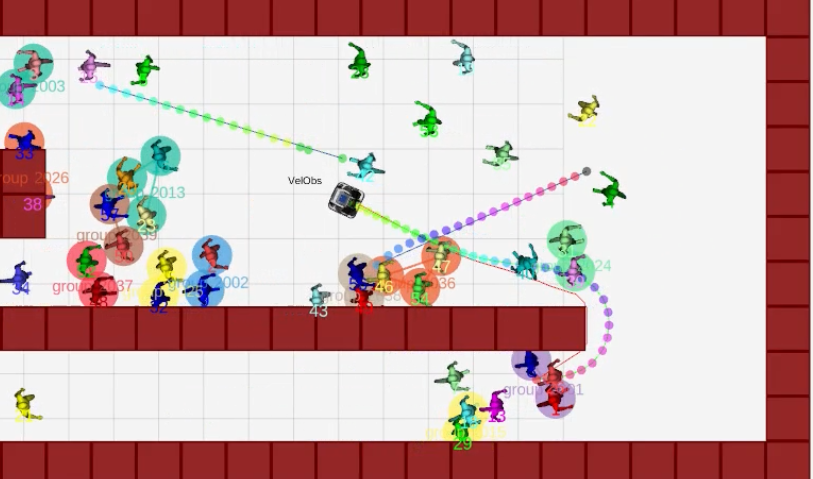}
      \caption{PedSim}
    \end{subfigure}
    \begin{subfigure}{.47\columnwidth}
      \includegraphics[width=\linewidth]{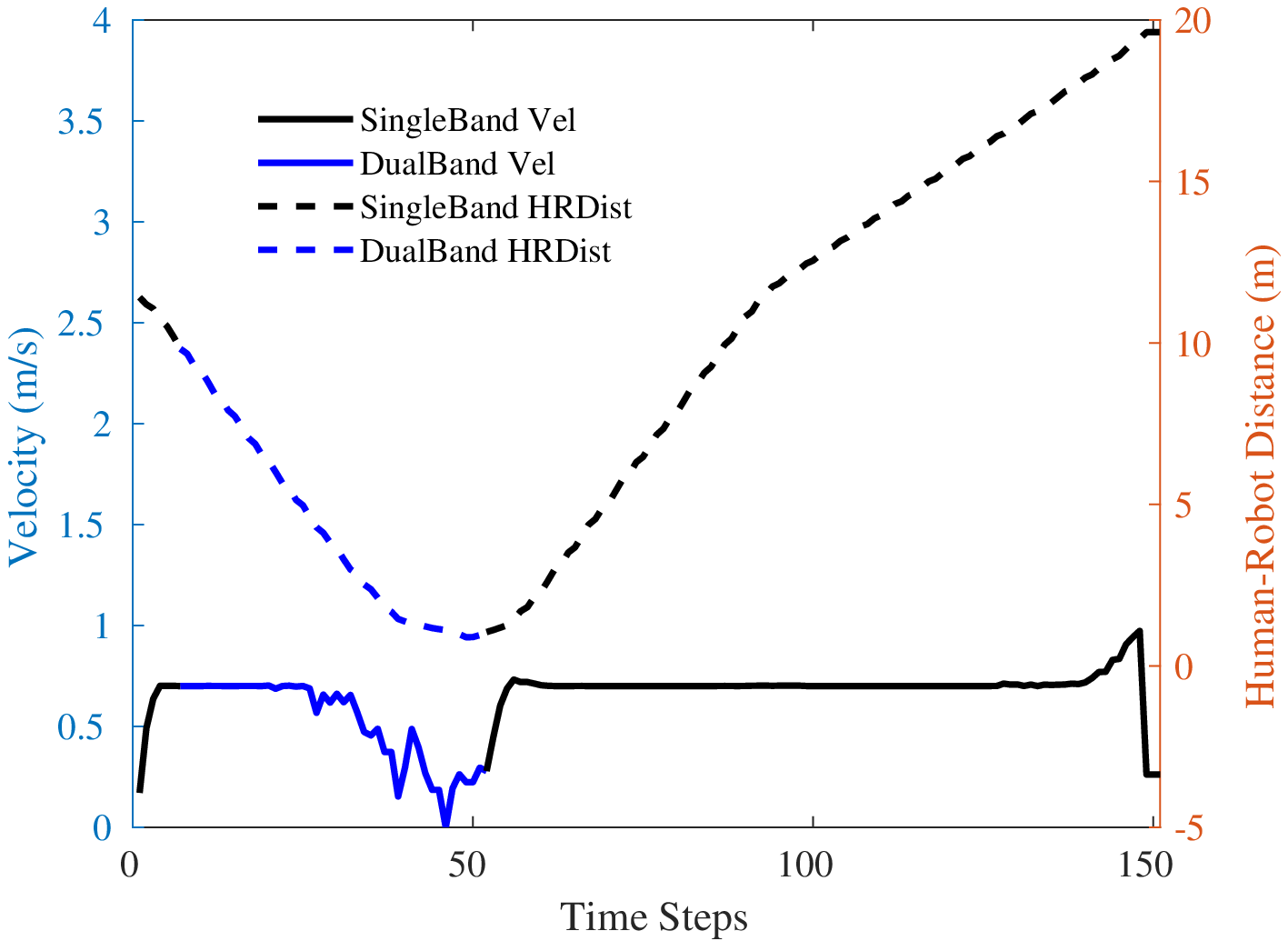}\caption{Pillar Corridor}
    \end{subfigure}
    \hspace{-0.5cm}
    \begin{subfigure}{.47\columnwidth}
      \includegraphics[width=\linewidth]{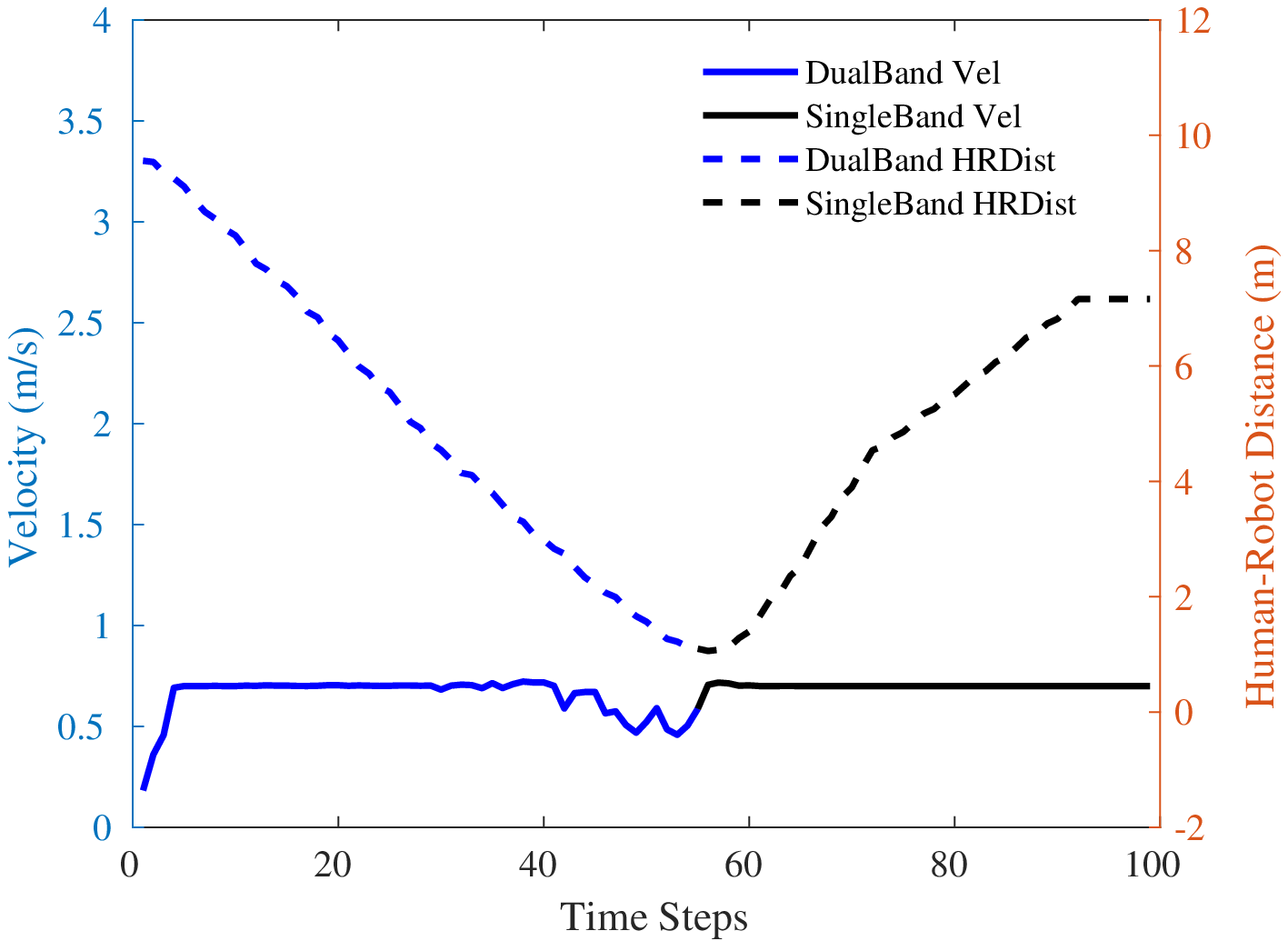}
      \caption{Wide Corridor}
    \end{subfigure}
    \hspace{-0.5cm}
    \begin{subfigure}{.47\columnwidth}
      \includegraphics[width=\linewidth]{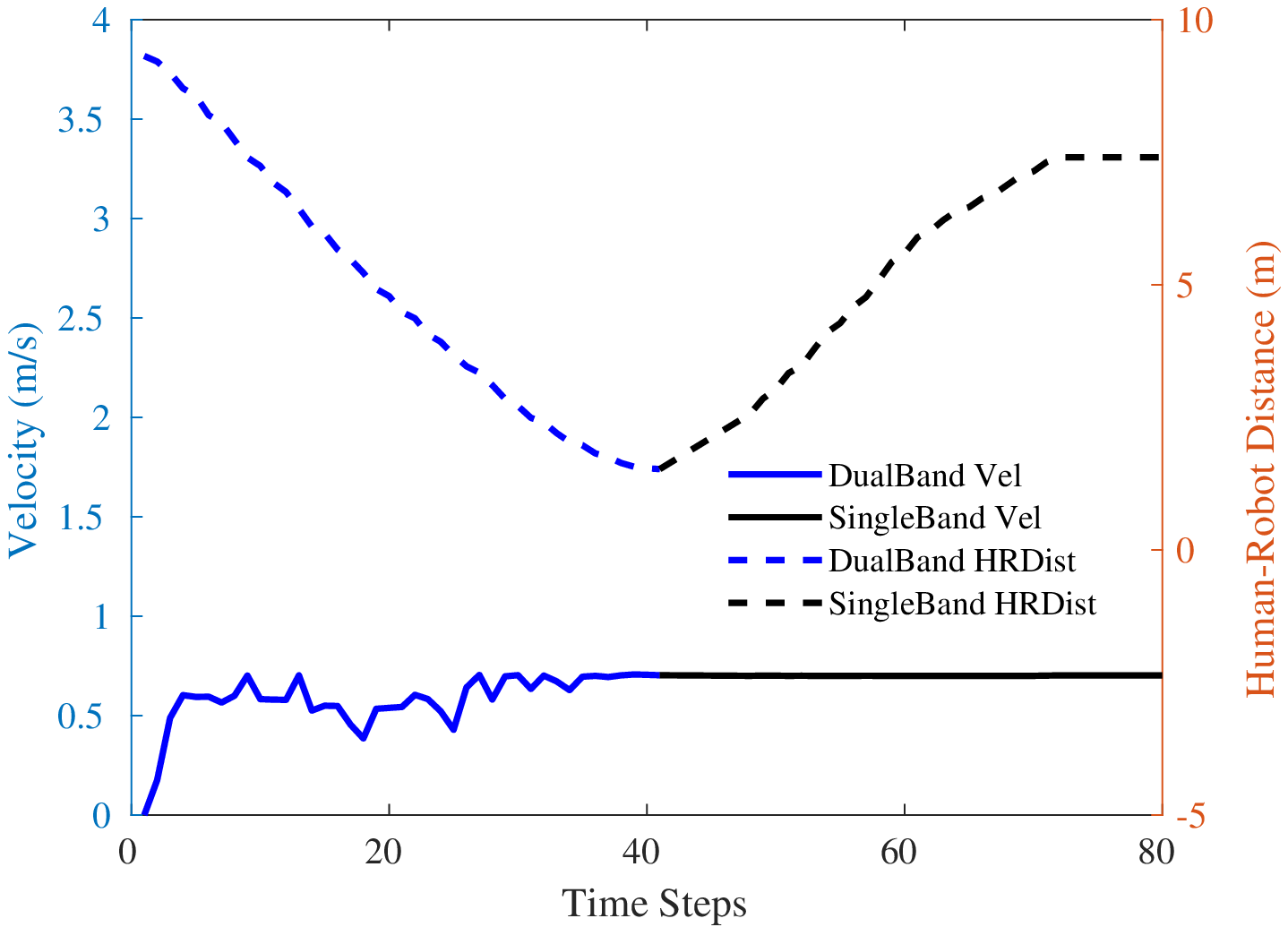}
      \caption{Open Space}
    \end{subfigure}
    \caption{(a) The robot running the proposed navigation system in the PedSim ROS pedestrian simulator. The robots planned trajectory and the predicted trajectories of the two nearest humans in \textit{VelObs} mode are shown. (b) A corridor with pillars, wide enough for only one agent at the side of pillar. (c) A wide corridor where the two agents have enough space to cross each other maintaining safe distances. (d) An open space scenario where robot have enough space to avoid and show its intention to the human well in advance. In (b), (c) and (d), the plots of robot's velocity and human-robot distance over time steps are shown below the scenario.}
    \label{fig:other_scenes}
    \vspace{-0.4cm}
\end{figure*}
This scenario occurs when a long corridor has to be traversed by two humans in opposite directions, and the corridor is wide enough only for a single human. In this case, one of them has to go back and wait for the other to cross. When one of the agents in this scenario is a robot, it becomes a little more complicated as the robot should back-off giving priority to the human while taking legible actions. Most of the existing planners either re-plan a long deviation to reach the goal or fail in this complicated situation. A more natural way to handle this would be to clear the way for the human and wait until the human crosses the robot to resume its goal. The \textit{Backoff-recovery} mode of our system does exactly this. To make the actions more legible, the robot moves back slowly without showing its back until it can go either left or right to clear the way. The snapshots from the simulated version of this scenario are shown in Fig. \ref{fig:narrow}. Each picture also shows the planned trajectory of the robot in each setting with the planning mode shown behind the robot. This scenario uses the \textit{PredictGoal} human path prediction, and the goal of the robot is on the other side of the corridor. Fig. \ref{fig:narrow} (a), shows the initial situation when the two agents are entering the narrow corridor. As the robot can see the human and the human is moving, it operates in \textit{DualBand} mode until the human blocks it's way completely. The human agent in this setting is controlled by the human simulator mentioned earlier. As soon as the robot finds itself blocked, it switches to \textit{VelObs} mode and checks for a possible solution. However, when it cannot find the solution after repeated checks, it switches to \textit{Backoff-recovery} mode after few seconds ($> 10s$) as shown in Fig. \ref{fig:narrow} (b). Fig. \ref{fig:narrow} (c) shows the robot waiting for the human to cross the corridor before it can resume its goal. The robot switches to \textit{SingleBand} mode and resumes its navigation to the goal as in Fig. \ref{fig:narrow} (d).

The plots of robot velocity and human-robot distance with respect to time steps for this scenario is shown in Fig. \ref{fig:narrow_plot}. As the human-robot distance decreases after a certain threshold, the velocity of the robot decreases like in the door crossing scenario (blue part). When the robot switches its mode from \textit{DualBand} to \textit{VelObs}, the robot tries to move in different directions causing the oscillations seen in the plot (red part). In the \textit{Backoff-recovery} mode, it maintains a constant velocity (green part) and stops waiting for the human. The human agent of the human simulator starts moving towards the robot as soon as it starts moving back. This explains a near-constant human-robot distance trend in green. Once the human passes the robot, it resumes its navigation in \textit{SingleBand} mode (black part).
\subsection{Results in the Crowd context and other scenarios}
We further tested our system in various scenarios, including crowds. For the simulation of crowds, we have used the PedSim ROS simulator and use the system in \textit{VelObs} mode. Fig. \ref{fig:other_scenes} (a) shows two snapshots from the tests.  The robot adds elastic bands to two of the nearest humans in the environment and successfully navigates the crowd generated by the simulator (shown in video). Further, it can be seen that the robot proactively clears the way for PedSim agents while navigating in the corridor shown, Fig. \ref{fig:other_scenes} (a).

We have simulated three other scenarios in MORSE: (1) Pillar Corridor, (2) Wide Corridor and (3) Open Space, and used our system to navigate them. These scenarios are shown in Fig. \ref{fig:other_scenes} (b), (c) and (d). In all three scenarios, human and robot goals are behind the other agent. In the Wide Corridor scenario, the robot predicts that the human's goal is behind its initial position and plans the human's trajectory. We use this planned human trajectory to control the human in this case, and hence it represents the ideal scenario for the planner. The scenario and its corresponding velocity and human-robot distance plots are shown in Fig. \ref{fig:other_scenes} (c). For the other two scenarios, the human agent was controlled by the human simulator, and the system uses \textit{PredictGoal} human path prediction. The velocity and human-robot distance plots for the Open Space (\ref{fig:other_scenes} (b)) scenario are similar to the Wide Corridor one (\ref{fig:other_scenes} (d)), however, in the Pillar Corridor (\ref{fig:other_scenes} (c)) scenario, the velocity almost reaches zero when the human is at the closest. This occurs as the robot has to wait behind the pillar for the human to cross. 
\subsection{Quantitative Results}
\begin{table*}[ht!]
    \centering
    \begin{tabular}{|c|c|c|c|c|c|c|}
    \hline
     & \multicolumn{3}{c|}{HATEB} & \multicolumn{3}{c|}{TPF}\\
    \cline{2-7}
    Experiment & \textit{Path Length} $(m)$ & \textit{Total Time} $(s)$ & \textit{Min  HRDist} $(m)$&  \textit{Path Length} $(m)$ & \textit{Total Time} $(s)$ & \textit{Min  HRDist} $(m)$\\
    \hline
    \textit{Open Space} & 9.2329 & \textbf{16.0117} & \textbf{1.2927} & \textbf{8.7892} & 24.7683 &0.9508\\
    \hline
    \textit{Narrow Passage} & 9.7279 & \textbf{17.8017} & 0.7061 & \textbf{9.4017} & 27.1633 & \textbf{0.935}\\
    \hline
    \textit{Pillar Corridor} & 19.1127 & \textbf{31.46} & \textbf{0.8948} & \textbf{18.4489} & 52.9417 & 0.7589\\
    \hline
    \textit{Narrow Corridor} & \textbf{23.5369} & \textbf{48.3850} & \textbf{0.6636} & - & - & -\\
    \hline
    \end{tabular}
    \caption{Mean values of the metrics over 10 repetitions in four different contexts. TPF failed in the Narrow Corridor case.}
    \label{results}
\end{table*}
In order to check the repeatability of our system and evaluate its performance with respect to the existing human-aware navigation planners, we selected four different simulated scenarios and repeated the same experiment 10 times in each of the scenarios. The scenarios we considered here include Open Space, Narrow Passage (similar to the one in \cite{singamaneni2020hateb}), Pillar Corridor and Narrow Corridor. The human in these scenarios is controlled by the improved human simulator mentioned earlier. In all these scenarios, our system produced consistent results over the repetitions with similar paths. Further, we compared it with the human-aware navigation planner presented in \cite{kollmitz_time_2015} that was designed for indoor home context. This system uses \textit{Timed Path Follower} (TPF) as its local planner, and its trajectory is highly dependant on the path produced by its global planner, \textit{Lattice Planner}. Note that our comparison was limited to only one planner as not many human-aware planners are available openly for indoor contexts. This planner was also made to run repeatedly 10 times in the above four scenarios. However, in the Narrow Corridor case, this planner failed to complete the navigation, and the robot got stuck in front of the human as the \textit{Lattice Planner} could not find a path.

We used three different metrics to present the comparisons between these two planners, the total length of the path taken, the total time to complete the scenario and the minimum human to robot distance that the planner encountered while executing each scenario. The average over the 10 runs is taken and presented in Table. \ref{results}. Note that the total time taken by the TPF is greater as its linear velocity was limited by the planner even though the same velocity and acceleration limits of the robot were provided to both the planners. In terms of the total path length, TPF always followed a lesser distance compared to \textit{HATEB}. This is because \textit{HATEB} took the larger deviations to either show intentions (in Open Space) or a clear path for the human (Pillar Corridor and Narrow Passage). However, in the Narrow Passage case, TPF produced better behaviour by waiting for the human to cross, while \textit{HATEB} blocked the way for a bit before clearing the way for the human. This can also be seen by comparing the minimum human-robot distance in this case. Finally, in terms of the minimum human-robot distances, \textit{HATEB} varies widely, as it handles each case differently. If the space is available, it takes a greater distance than TPF, otherwise, it slows down the robot's velocity and approaches a little closer to the human. In TPF, this metric produces similar results in two of the three scenarios. In the Pillar Corridor, this metric has a lesser value compared to \textit{HATEB} as the robot goes towards the wall opposite to pillar and waits, instead of going behind the pillar.      
\section{Experiments with real humans} \label{sec_4}
We deployed our system on a real robot platform, Pepper\footnote{\url{https://www.ald.softbankrobotics.com/en/pepper}}, for some real-world experiments in our lab. For the human detection and tracking, we used the OptiTrack\footnote{\url{http://www.optitrack.com/}} motion capture system that publishes the positions and velocities of the tracked humans at 10 Hz. We present results from two experiments, one in open space and the other in the narrow corridor. \begin{figure}[ht!]
    \centering
    \begin{subfigure}{0.6\columnwidth}
      \includegraphics[width=\linewidth]{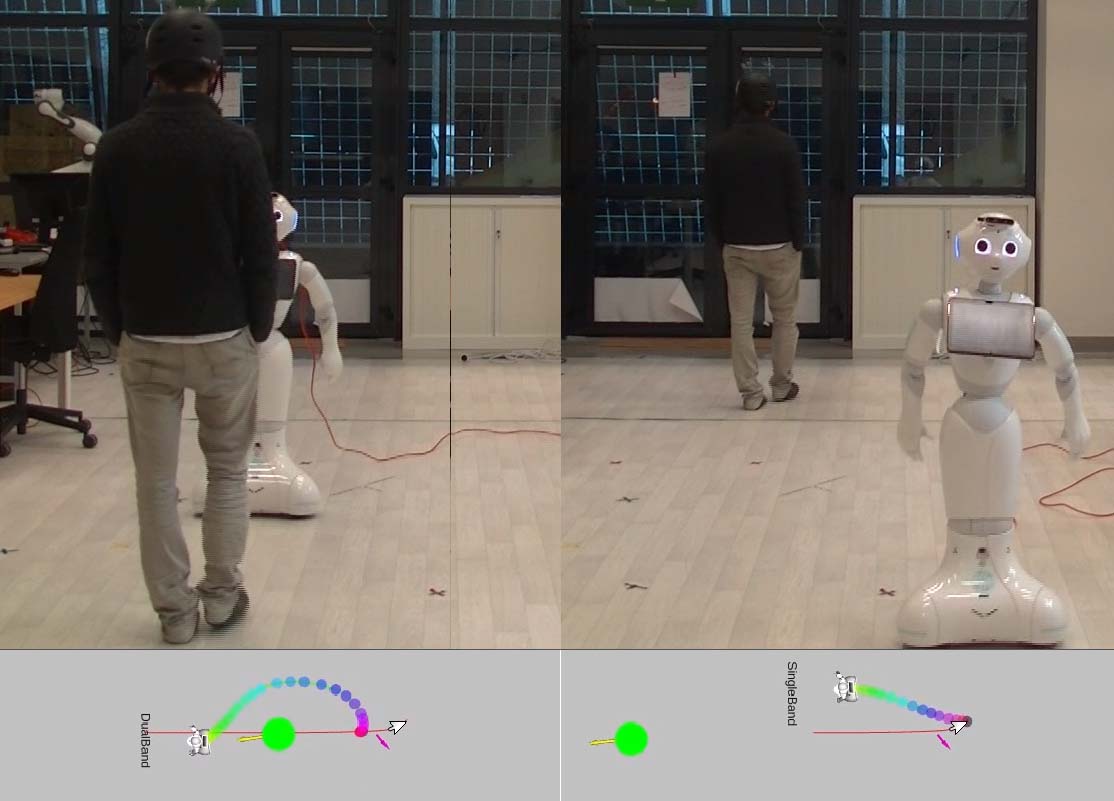}
    \end{subfigure}
      \begin{subfigure}{0.6\columnwidth}
      \includegraphics[width=\linewidth]{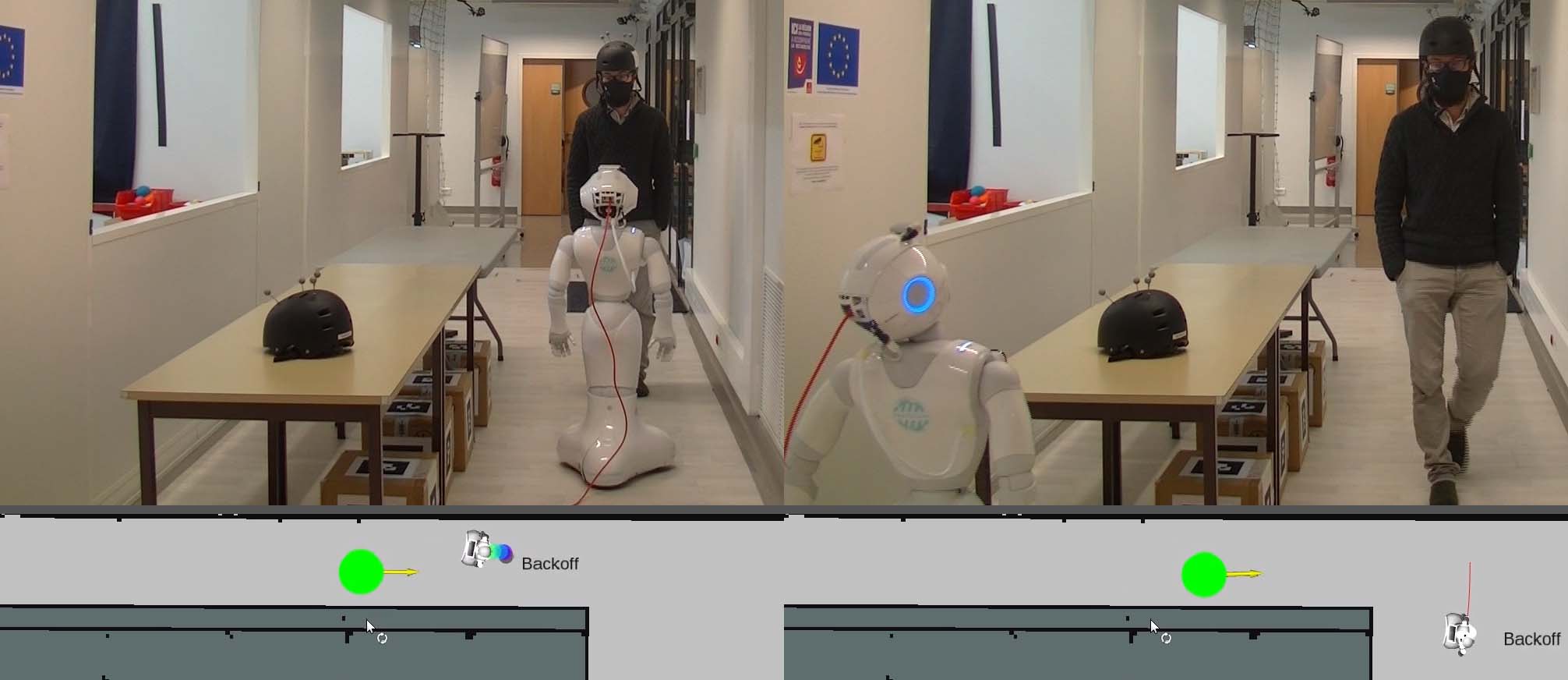}
    \end{subfigure}
    \caption{Our system deployed and tested on a real robot, Pepper. The top pictures show the open space scenario, and the bottom pictures show the test of the \textit{Backoff-recovery} mode.}
    \label{fig:real_exp}
\end{figure}
Fig. \ref{fig:real_exp} shows the instances from these experiments. The pictures on the top show the open space scenario with the planned trajectory of the robot shown at the bottom of each instance. In this scenario, the human approaches very close to the robot at the crossing point, and this makes the robot slow down, back off a little and then re-plan its trajectory (please see the video\footnote{Video Link: \url{https://youtu.be/Jwi_gYva_VQ}} attachment). The bottom part of Fig. \ref{fig:real_exp} shows instants from the test of \textit{Backoff-recovery} mode in our system. The human stands in the corridor blocking the robot's way. The planner starts in \textit{DualBand} mode and finally switches to \textit{Backoff-recovery} mode as there is no solution. The robot slowly backs off and clears the way for the human by moving to the left side of the corridor, as seen in Fig. \ref{fig:real_exp}.

\section{Discussion and Conclusion} \label{sec_5}
In this work, we proposed a new human-aware navigation planner that can handle a variety of human-robot contexts. It was able to handle both outdoor crowd scenario and indoor intricate scenarios, thanks to the different planning modes and tunable parameters in the system. These planning modes are at the control level and hence differs from higher-level modes as used in \cite{mehta_autonomous_2016}. Consider the \textit{Backoff-recovery} mode for example, instead of going into the corridor, the robot could have stopped (or back-off) as soon as it sees a person and waited for the human to complete his/her navigation. Nonetheless, this may not be possible in very long corridors due to the lack of visibility. By employing a higher level planner, this case could be handled much more efficiently, but our focus is on providing this feature at the control level. We introduced \textit{Human Safety} and \textit{Human Visibility} layers into the system through costmaps to address the static human scenarios. For handling the dynamic humans, we have used a variety of human-aware constraints in \textit{HATEB} along with visibility and planning radius. The proposed system also provided different types of human path prediction methods. We introduced two new human-aware constraints in addition to the previous ones present in \textit{HATEB} to offer a more legible trajectory. Further, we evaluated our system in a variety of simulated scenarios and presented both qualitative and quantitative results. Finally, the real-world tests on a robotic platform were presented.

\textit{Limitations and Future Work}: One of the major limitations of our system could be computational complexity as it performs optimization in each control loop. However, it does not affect the real-time performance of the robot in \textit{SingleBand} and \textit{Backoff-recovery} modes (10Hz). In the other modes, it may lead to a little reduced control rate (8-9 Hz), however still in realtime. One of the immediate future works could be to use a higher-level planner on the top of our system and include contexts like following or accompanying a human. Currently, the system is not designed for handling groups of people differently, and we plan to include it in the future version of the system. One more possible future path would be to include human detection and tracking system along with the navigation system.

\bibliographystyle{unsrt}
\bibliography{references}

\end{document}